\documentclass[lettersize,conference]{ieeeconf}
\IEEEoverridecommandlockouts

\usepackage{times}
\usepackage{amsmath}
\usepackage{amssymb}
\usepackage{amsfonts}
\usepackage{xcolor}
\usepackage{bm}
\usepackage{graphicx}
\usepackage{float}
\usepackage{multirow}
\usepackage{booktabs}
\usepackage{url}
\usepackage{verbatim}
\usepackage{array}
\usepackage[caption=false,font=normalsize,labelfont=sf,textfont=sf]{subfig}
\usepackage{stfloats}
\usepackage{textcomp} 

\usepackage{algorithm}
\usepackage{algorithmic}

\def\BibTeX{{\rm B\kern-.05em{\sc i\kern-.025em b}\kern-.08em
    T\kern-.1667em\lower.7ex\hbox{E}\kern-.125emX}}

\makeatletter
\newcommand\fs@norules{\def\@fs@cfont{\bfseries}\let\@fs@capt\floatc@ruled
  \def\@fs@pre{}%
  \def\@fs@post{}%
  \def\@fs@mid{\kern3pt}%
  \let\@fs@iftopcapt\iftrue}
\makeatother
\floatstyle{norules}
\restylefloat{algorithm}

\title{\LARGE \bf
Integrating Ergonomics and Manipulability for Upper Limb Postural Optimization in Bimanual Human-Robot Collaboration
}

\author{Chenzui Li, Yiming Chen, Xi Wu, Giacinto Barresi, and Fei Chen%
\thanks{This work was supported by the Research Grants Council of the Hong Kong SAR under Grant 24209021, 14222722, 14211723 and C7100-22GF and the InnoHK initiative of the Innovation and Technology Commission of the Hong Kong Special Administrative Region Government via the Hong Kong Centre for Logistics Robotics. (\textsuperscript{\textdagger}\,Corresponding author: Fei Chen)}%
\thanks{Chenzui Li, Yiming Chen, Xi Wu, and Fei Chen are with the Department of Mechanical and Automation Engineering, T-Stone Robotics Institute, The Chinese University of Hong Kong, Hong Kong (e-mail: \{czli, ymchen, xwu\}@mae.cuhk.edu.hk; f.chen@ieee.org).}%
\thanks{Giacinto Barresi is with the Bristol Robotics Laboratory, UWE Bristol, England, UK (e-mail: Giacinto.Barresi@uwe.ac.uk).}%
\thanks{Fei Chen is also with the HKCLR, Hong Kong.}
}

\begin{document}
\maketitle

\begin{abstract}
This paper introduces an upper limb postural optimization method for enhancing physical ergonomics and force manipulability during bimanual human-robot co-carrying tasks. Existing research typically emphasizes human safety or manipulative efficiency, whereas our proposed method uniquely integrates both aspects to strengthen collaboration across diverse conditions (e.g., different grasping postures of humans and different shapes of objects). Specifically, the joint angles of a simplified human skeleton model are optimized by minimizing a cost function to prioritize safety and manipulative capability. To guide humans toward the optimized posture, the reference end-effector poses of the robot are generated through a transformation module. A bimanual model predictive impedance controller (MPIC) is proposed for our human-like robot, CURI, to recalibrate the end-effector poses through planned trajectories. The proposed method has been validated with various subjects and objects during human-human collaboration (HHC) and human-robot collaboration (HRC). The experimental results demonstrate significant improvement in muscle conditions by comparing the activation of target muscles before and after optimization.
\end{abstract}

\section{Introduction} \label{introduction}

The rapid integration of robots into collaborative tasks with humans has become increasingly prevalent in both industrial and domestic environments, driven by the need for efficiency and safety in handling heavy or cumbersome objects \cite{evrard2009teaching}. In scenarios such as manufacturing, logistics, and even domestics, the joint carrying of objects by humans and robots offers substantial advantages, including reduced physical strain on human workers and enhanced precision and stability in object manipulation. 

Current research faces significant challenges, primarily in designing efficient collaboration skills and addressing programming complexities in unstructured settings \cite{mukherjee2022survey}. The integration of collaborative robots (cobots) into various operational contexts has been facilitated by intuitive programming methods, which simplify their deployment in user-friendly ways. Besides, other methods like Learning from Demonstrations (LfD) depend heavily on demonstrated trajectories and idealized postures during interactions, which may not be suitable for partners of varying heights or preferences \cite{nemec2018human}. In real-world co-carrying tasks, while bimanual co-carrying is commonly employed, it is infrequently addressed in research due to its inherent complexity. Another issue is that humans tend to adopt their postures based on environmental constraints and the initial object poses, which unconsciously leads to postural inefficiencies and potential long-term health issues like musculoskeletal disorders (MSD) \cite{da2010risk}. These variations highlight the need for more adaptable approaches in HRC to accommodate real-world variability and ensure human safety.

\begin{figure}[!t]
\centering
\includegraphics[width=1\linewidth]{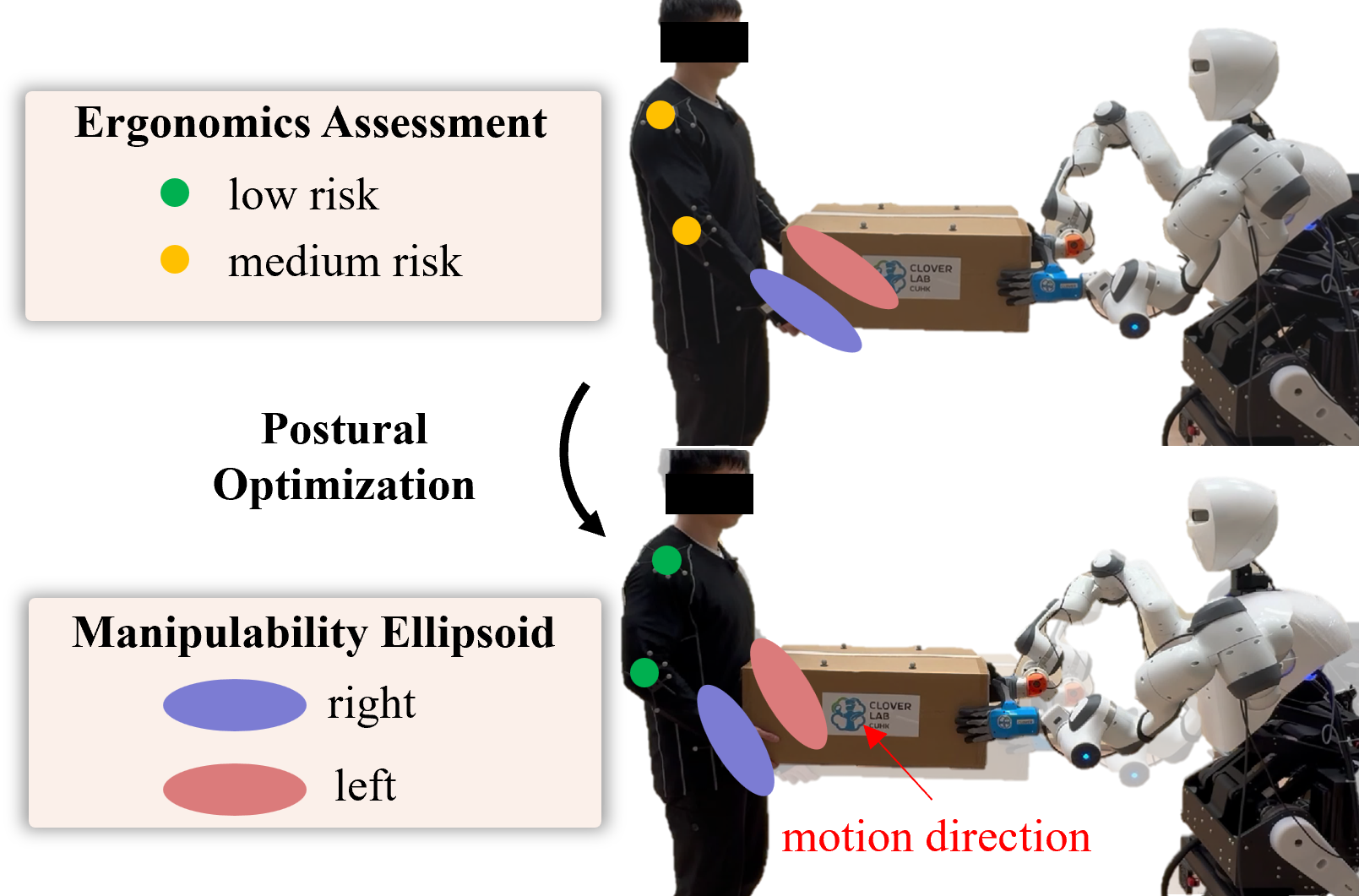} 
\caption{Illustration of upper limb postural optimization based on ergonomics and manipulability in human-robot co-carrying tasks.}
\label{Fig.cover}
\vspace{-4mm}
\end{figure}

Ergonomics studies the interactions between humans and other elements of a system to improve the human well-being and the overall system performance. Among its main branches, physical ergonomics aims to design workspaces, tools, and tasks to align with human body characteristics and movements. Cobots have broadened their application to improve physical ergonomics, especially for postural correction \cite{cardoso2021ergonomics}. Ergonomic-basd optimization methods have been proposed whether offline \cite{van2020predicting} or online \cite{shafti2019real} for HRC. An ergo-interactive framework is raised by integrating virtual ergonomic forces into learned Riemannian dynamic movement primitives to promote ergonomic human postures during the co-carrying of objects \cite{liao2023ergo}. In \cite{van2020predicting}, the Rapid Entire Body Assessment (REBA) method is applied to estimate physical ergonomics while the cobot adjusts human postures with an informed graph search algorithm. Another REBA-based ergonomic optimization framework which uses virtual elements to model and optimize human posture in real-time and a robot-controlled workpiece positioning system to adjust the human posture during tasks is proposed \cite{el2022virtual}.

However, existing ergonomic-based approaches always prioritize static and oversimplified task executions, neglecting the dynamic nature of human preferences and the complexities of real-world environments (e.g., diverse shapes of objects). When handling large or heavy objects, it is essential to ensure that the individual's posture is properly aligned to support the load effectively. From an ergonomic perspective, the ability to exert sufficient force (e.g., manipulability \cite{raessa2020human}) to manage the load and facilitate movements is equally critical. To tackle the problem, a novel metric is proposed which integrates physical ergonomics and muscular manipulability into a quality distribution for both pre-computation and real-time application during a unimanual collaborative task \cite{figueredo2020human}. Nevertheless, extending this to bimanual tasks will introduce additional challenges such as dual-arm coordination.


To address these limitations, this paper proposes a novel upper-limb postural optimization method aimed at simultaneously improving physical ergonomics and force manipulability for human partners in bimanual co-carrying tasks with cobots. The key contributions are summarized as follows:
\begin{itemize}
  \item [1)]
  A unique integration of ergonomics and manipulative capacities has been raised for human postural optimization, tailored specifically for the complexities of bimanual tasks in HRC settings.
  \item [2)]
  A model predictive impedance controller (MPIC) has been raised with bimanual coordination, enabling dynamic cobot adjustment while enforcing state constraints and improving perturbation resistance.
  \item [3)]
  Multiple human subjects have experimentally validated the framework by co-carrying diverse objects through HHC and HRC with diverse initial dual-arm grasping postures. The effectiveness is proved by comparing the muscle activation levels before and after optimization. 
\end{itemize}

\begin{figure*}[!t]
\centering
\includegraphics[width=1\linewidth]{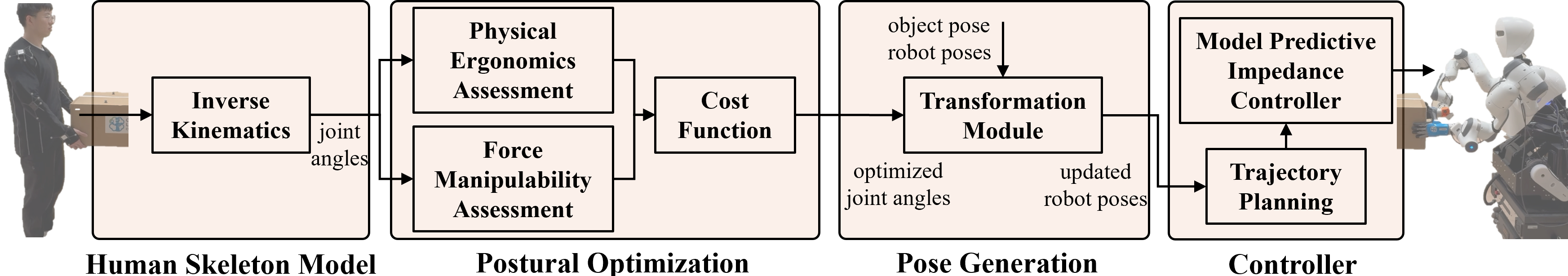} 
\caption{Illustration of the proposed collaborative framework for Human-Robot Co-Carrying. \textit{Human Skeleton Model:} A simplified human upper limb kinematic model is given to calculate specific joint angles. \textit{Postural Optimization:} An optimization method is proposed to enhance human posture by integrating ergonomics and manipulability. \textit{Pose Generation:} The robot desired end effector poses are generated based on a transformation module. \textit{Controller:} An MPIC is applied for our human-like robot, CURI, to execute the planned trajectories.}
\label{Fig.framework} \vspace{-3mm}
\end{figure*}

\section{Methodology} \label{method}
The overall framework (see Fig. \ref{Fig.framework}) involves four main components: human skeleton model, postural optimization, pose generation, and controller. Initially, a simplified human skeleton model is applied for real-time joint angle calculation through inverse kinematics, which can assess the physical ergonomics and the force manipulability of the partners. The postural optimization module further introduces a self-defined cost function and minimizes it with an optimization method to get the optimized joint angles. Then a transformation module is proposed to generate reference robot poses according to initial and optimized human joint angles, the object pose, and the initial robot poses. Finally, the minimum jerk trajectory planning is conducted from initial poses to reference poses and an MPIC is applied to execute bimanual movement. The details are explained in the following subsections.

\subsection{Modeling}
Ergonomics and manipulability are involved in our proposed optimization method. Since humans participate in bimanual co-carrying tasks, the upper limbs are considered for analysis. 

\subsubsection{Human Upper Limb Kinematic Model}\label{model}
A simplified kinematic model \cite{klopvcar2005kinematic} for determination of interrelations between coordinates is given in Fig. \ref{Fig.skeleton}. $\boldsymbol{R}_1$ is associated with the shoulder abduction/adduction movement, $\boldsymbol{R}_2$ with the shoulder flexion/extension movement, $\boldsymbol{R}_3$ with the internal/external shoulder rotation, and $\boldsymbol{R}_4$ represents the elbow flexion-extension. The coordinates of rotations $\boldsymbol{R}_1$, $\boldsymbol{R}_2$, $\boldsymbol{R}_3$, $\boldsymbol{R}_4$ are joint angles $q_1$, $q_2$, $q_3$, $q_4$, respectively. Parameter $\boldsymbol{d}_{ua}$ represents the humerus
and is the length between the outer shoulder joint and the elbow joint, $\boldsymbol{d}_{fa}$ represents the forearm and is the length between the elbow joint and wrist joint $\boldsymbol{p}_w$. The reference coordinate frame is attached to the torso in the center of the inner shoulder joint as shown in Fig. \ref{Fig.skeleton}. In the reference pose of the arm, when all joint coordinates are zero, $\boldsymbol{q} = [q_1,..., q_4]^T = 0$, $\boldsymbol{R}_1$ is parallel to $y$, $\boldsymbol{R}_2$ is parallel to $x$, $\boldsymbol{R}_3$ is parallel to $z$, and $\boldsymbol{R}_4$ is parallel to $x$, while the upper arm link ($\boldsymbol{d}_{ua}$) is parallel to $z$, and the forearm link ($\boldsymbol{d}_{fa}$) is parallel to axis $y$.

Accordingly, the position of wrist point $\boldsymbol{p}_w$ with respect to the reference coordinate frame can be calculated by:

\begin{equation}\label{eq1}
  \boldsymbol{p}_w = \boldsymbol{R}_1 \boldsymbol{R}_2 \boldsymbol{R}_3 (\boldsymbol{d}_{ua} + \boldsymbol{R}_4 \boldsymbol{d}_{fa}),
\end{equation}
where $\boldsymbol{d}_{ua} = [0, 0, -d_{ua}]^T$ and $\boldsymbol{d}_{fa} = [0, d_{fa}, 0]^T$. $\boldsymbol{R}_{i}$ are the following rotation matrices:

\begin{equation}
\begin{aligned}
\boldsymbol{R}_{1} = \left [ \begin{matrix}
    c_1 & 0 & s_1 \\
    0 & 1 & 0 \\
    -s_1 & 0 & c_1
\end{matrix} \right ] , \quad
\boldsymbol{R}_{2} = \left [ \begin{matrix}
    1 & 0 & 0 \\
    0 & c_2 & -s_2 \\
    0 & s_2 & c_2
\end{matrix} \right ] , \\
\boldsymbol{R}_{3} = \left [ \begin{matrix}
    c_2 & -s_2 & 0 \\
    s_2 & c_2 & 0 \\
    0 & 0 & 1
\end{matrix} \right ] , \quad
\boldsymbol{R}_{4} = \left [ \begin{matrix}
    1 & 0 & 0 \\
    0 & c_4 & -s_4 \\
    0 & s_4 & c_4
\end{matrix} \right ] .
\end{aligned}
\end{equation}

\subsubsection{Manipulability Computation}
Velocity/force manipulability ellipsoids are kinetostatic performance measurements and indicate the capability to generate velocity/force in different directions at a given joint configuration \cite{bicchi1995mobility}. Specifically, the velocity manipulability ellipsoid is a geometric representation that characterizes the feasible Cartesian motions, given all possible unit norm joint velocities. In our case, for a $4$-DOF model, the velocity manipulability can be quantified through the relationship between the task velocities $\boldsymbol{\dot{x}}$ and the joint velocities $\boldsymbol{\dot{q}}$. This relationship is mathematically expressed by the equation:

\begin{equation}\label{eq3}
\boldsymbol{\dot{x}} = \boldsymbol{J}(\boldsymbol{q})\boldsymbol{\dot{q}},
\end{equation}
where $\boldsymbol{q} \in \mathbb{R}^n$ represents the vector of joint positions, and $\boldsymbol{J} \in \mathbb{R}^{6 \times 4}$ denotes the Jacobian matrix. 

\begin{figure}[!t]
\centering
\includegraphics[width=1\linewidth]{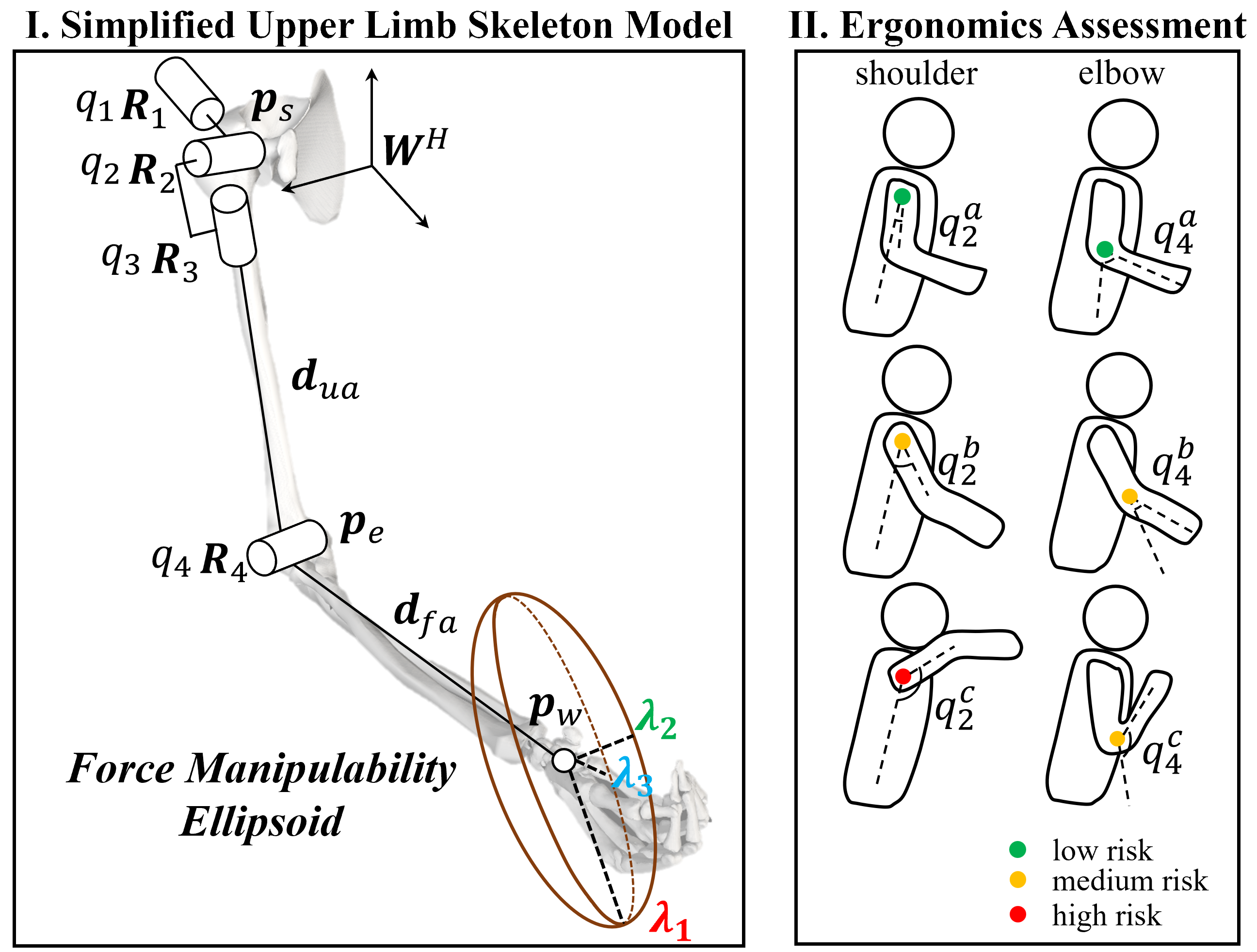} 
\caption{(I) Illustration of the simplified human right upper limb skeleton model and force manipulability ellipsoid, (II) examples of REBA physical ergonomic assessment with human upper limb joints (shoulder flexion/extension and elbow flexion/extension, respectively).}
\label{Fig.skeleton} \vspace{-3mm}
\end{figure}

In this paper, the position component of the force manipulability ellipsoid is used and it can be calculated based on the simplified upper limb model of the human shown in Fig. \ref{Fig.skeleton}.a. By monitoring the human joint positions including the shoulder, elbow, and wrist during the task execution, this model can compute the human upper limb joint configuration with inverse kinematics and further generate the human arm Jacobian $\boldsymbol{J} \in \mathbb{R}^{3 \times 4}$. The force manipulability ellipsoid is then computed as $\boldsymbol{M^F} = (\boldsymbol{J} \boldsymbol{J}^T)^{-1} \in \mathbb{R}^{3 \times 3}$ with eigenvalues due to the velocity-force duality \cite{yoshikawa1985manipulability}. 

\subsubsection{Ergonomics Assessment}
The REBA methodology \cite{hignett2000rapid} has been adopted for assessing the physical ergonomics in terms of upper limbs activities. Some modifications have been implemented to the original standardized ergonomics metric to better suit our method. However, it is worth noting that this metric can easily be replaced by any other assessment that utilizes data on human postures and carrying loads. The original REBA method employs a series of tables that assess human posture based on joint angles, supplemented by additional pertinent information. One example is given in Fig. \ref{Fig.skeleton}.b to show the evaluation of the shoulder and elbow ergonomics.

For the bimanual tasks we are considering, we evaluate the arm that is the least ergonomic in order to compute the worst-case score. Besides, the original REBA method employs discrete assessments according to the joint degrees, which can introduce significant challenges during optimization processes. Specifically, discrete indicators may lead to issues such as premature convergence and an increased likelihood of the optimization algorithms becoming trapped in local optima, thereby hindering the achievement of global solutions \cite{al2021discrete}. Continuous functions are applied to represent these indicators to mitigate this problem. This modification enhances the optimization efficiency by facilitating smoother gradients and more dynamic adjustment capabilities, ultimately reducing the risk of stagnation at suboptimal solutions and promoting robust convergence towards global optima.

\subsection{Postural Optimization}
The optimization method is introduced to enhance human postures regarding ergonomics and manipulability within shared workspaces (Fig. \ref{Fig.framework}). After assessing the upper limb ergonomic scores as well as force manipulability ellipsoids, a cost function is designed and the constraints of the optimization problem are given. This optimization problem aims to find the desired joint angles that make the human friendly while satisfying operability. 

Specifically, the forward kinematics can be written based on (\ref{eq1}) with $\boldsymbol{p}_{s}, \boldsymbol{p}_{e}, \boldsymbol{p}_{w}$ represent the positions of the human shoulder, elbow, and wrist in the reference coordinates, respectively.
\begin{equation}\label{eq4}
\begin{aligned}
\boldsymbol{p}_{e} &= \boldsymbol{p}_{s} + \boldsymbol{R}_1 \boldsymbol{R}_2 \boldsymbol{R}_3 \boldsymbol{d}_{ua}, \\
\boldsymbol{p}_{w} &= \boldsymbol{p}_{s} + \boldsymbol{R}_1 \boldsymbol{R}_2 \boldsymbol{R}_3 (\boldsymbol{d}_{ua} + \boldsymbol{R}_4 \boldsymbol{d}_{fa}),
\end{aligned}
\end{equation}

During co-carrying tasks, the human joint positions are recorded online and the joint angles $\boldsymbol{q}$ can be calculated based on the inverse kinematics problem:
\begin{equation}
\begin{aligned}
&\min_{\boldsymbol{q}} \left( \left\| \boldsymbol{p}_{e\_d} - \boldsymbol{p}_{e} \right\|^2 + \left\| \boldsymbol{p}_{w\_d} - \boldsymbol{p}_{w}) \right\|^2 \right) , \\
\text{s.t.} &\quad -\pi/18 \le q_1 \le 17\pi/18 \\
&\quad -\pi/3 \le q_2 \le 17\pi/18 \\
&\quad -\pi/3 \le q_3 \le \pi/2 \\
&\quad -\pi/2 \le q_4 \le \pi/3
\end{aligned}
\end{equation}

With these joint angles, the Jacobian of human upper limb $\boldsymbol{J}$ and the arm force manipulability ellipsoid $\boldsymbol{M^F}$ is then computed according to (\ref{eq3}). The major axis of the ellipsoid aligned to the eigenvector associated with the maximum eigenvalue $\lambda^{\boldsymbol{M^F}}_{max}$ indicates the direction where the greater force can be generated. In this paper, the force manipulability ellipsoid is applied to ensure the human has sufficient manipulability in the direction of movement and load. 

The assessment of upper limb ergonomics is then calculated based on the modified REBA method. The original method scores joint angles by assigning uniform values within a range. For example, when referring to shoulder flexion and extension (Fig. \ref{Fig.skeleton}), upper limb score plus 1 when $q_2 \in \left(-\frac{2 \pi}{9}, \frac{2 \pi}{9}\right)$, plus 2 when $q_2 \in \left(\frac{2 \pi}{9}, \frac{\pi}{4}\right)$, plus 3 when $q_2 \in \left(\frac{\pi}{4}, \frac{\pi}{2}\right)$, and plus 4 when $q_2 \in \left(\frac{\pi}{2}, \pi\right)$. We modify the discrete scoring mechanism to a continuous one, expressed as linear functions: 
\begin{equation}
\begin{aligned}
s_{s} = \left\{ 
\begin{array}{ll}
\left| q_2 \right| / \frac{2 \pi}{9} , & \text{if } q_2 \in \left(-\frac{2 \pi}{9}, \frac{2 \pi}{9}\right) \\
1 + \left| q_2 - \frac{2 \pi}{9} \right| / \frac{2 \pi}{9} , & \text{if } q_2 \in \left(\frac{2 \pi}{9}, \frac{\pi}{4}\right) \\
2 + \left| q_2 - \frac{\pi}{4} \right| / \frac{\pi}{4} , & \text{if } q_2 \in \left(\frac{\pi}{4}, \frac{\pi}{2}\right) \\
3 + \left| q_2 - \frac{\pi}{2} \right| / \frac{\pi}{2} , & \text{if } q_2 \in \left(\frac{\pi}{2}, \pi\right)
\end{array}
\right.
\end{aligned}
\end{equation}

After scoring all relevant joint angles of shoulder and elbow joints, the upper limb ergonomic score is defined as $s(\boldsymbol{q}_i) = s_{s}(\boldsymbol{q}_i) + s_{e}(\boldsymbol{q}_i), \ i=l, r$. Since bimanual collaboration is considered, $\boldsymbol{q} = [\boldsymbol{q}_l^T, \boldsymbol{q}_r^T]^T$, the overall ergonomic score is computed as:
\begin{equation}
s(\boldsymbol{q}) = \text{max}(s(\boldsymbol{q}_l), s(\boldsymbol{q}_r)),
\end{equation}

Hence, the optimization problem can be written as the following equations:
\begin{equation}
\begin{aligned}
&\min_{\boldsymbol{q}} (\alpha (s(\boldsymbol{q}))^2 + \beta (m(\boldsymbol{q}))^2 + \gamma \sum_{i=1}^n (\boldsymbol{q}_i- \boldsymbol{q}_{i\_init})^2), \\
\text{s.t.} &\quad \left\| \boldsymbol{x}^{w}_{l_\_new} - \boldsymbol{x}^{w}_{r_\_new} \right\| - \left\| \boldsymbol{x}^{w}_l - \boldsymbol{x}^{w}_r \right\| \le \epsilon\\
\end{aligned}
\end{equation}

The first item $s(\boldsymbol{q})$ is the value of ergonomics, the second item $m(\boldsymbol{q}) = \sqrt{ (m(\boldsymbol{q}_l) - m_0)^2 + (m(\boldsymbol{q}_r) - m_0)^2}$ is the value of manipulability where $m(\boldsymbol{q}_l)$ and $m(\boldsymbol{q}_r)$ are the lengths of the maximum eigenvalue $\lambda^{\boldsymbol{M^F}}_{max}$ along the load direction and $m_0$ the reference norm. The third item represents the penalty if large deviations of joint angles from initial angles occur. $\alpha$, $\beta$, and $\gamma$ represent the scaling factors for each component, respectively. Two constraints are considered including the relative pose constraint between the human wrists since the contact points are fixed and the joint angle range constraint, which is the same as the constraint for inverse kinematic optimization. 

After obtaining the optimized joint angles $\boldsymbol{q}_{opt}$, the optimized human wrist positions ($\boldsymbol{p}_{w\_l\_opt}^R$, $\boldsymbol{p}_{w\_r\_opt}^R$) can be computed according to (\ref{eq4}). 

\begin{figure}[!t]
\centering
\includegraphics[width=0.95\linewidth]{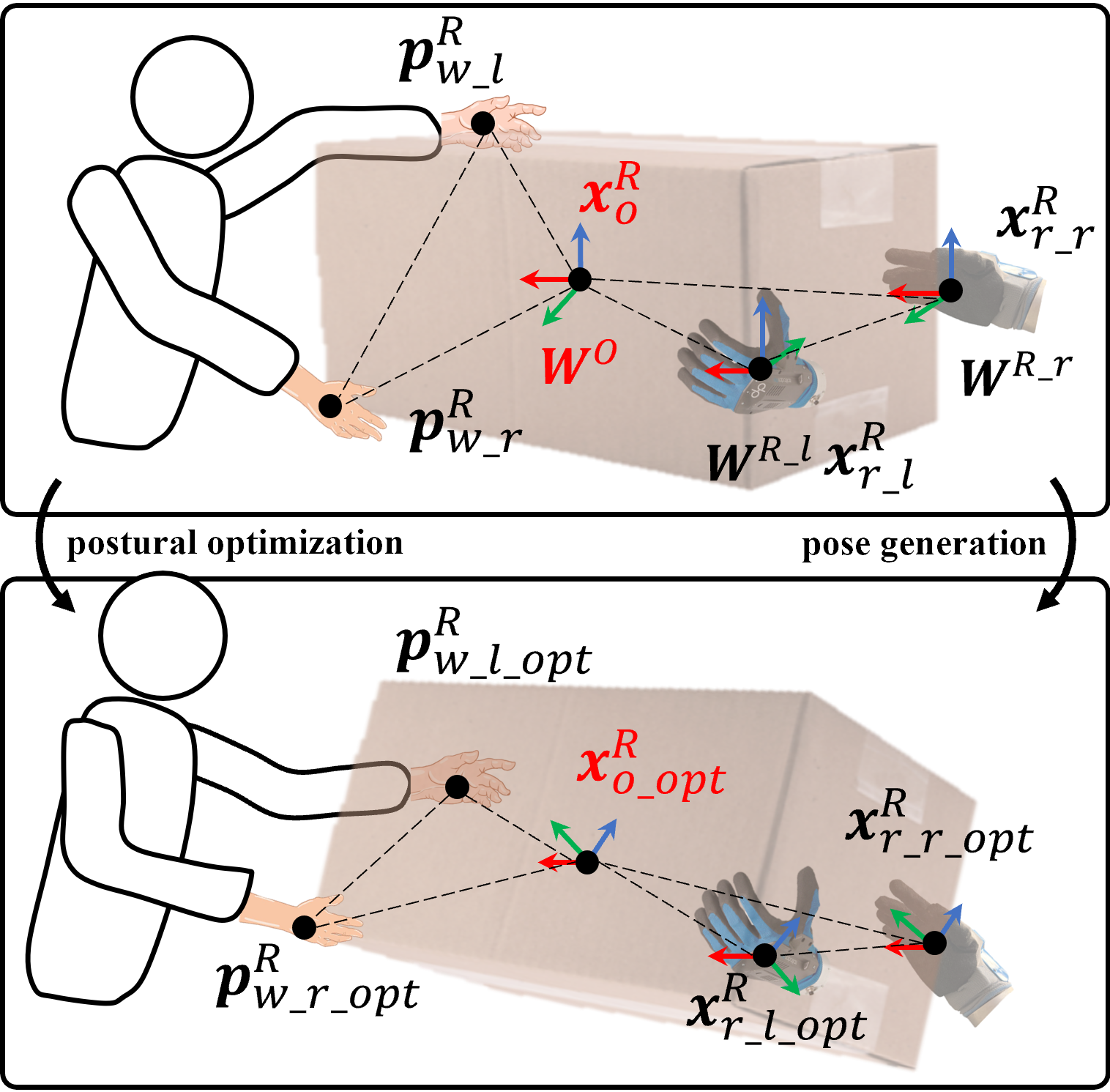} 
\caption{Diagram of pose generation to calculate the reference end effector poses of the cobot based on human initial and optimized wrist positions, object initial pose, and cobot initial end effector poses.}
\label{Fig.optimization} \vspace{-3mm}
\end{figure}

\subsection{Robot Pose Generation}


The end effector poses of the bimanual cobot will be generated based on the optimized joint angles. The updated object pose is first calculated. Since the contact points between the human and the object are fixed, then the corresponding object pose $\boldsymbol{x}_{o\_opt}^R$ can be given with respect to the robot frame based on the initial wrist positions ($\boldsymbol{p}_{w\_l}^R$, $\boldsymbol{p}_{w\_r}^R$), optimized wrist positions ($\boldsymbol{p}_{w\_l\_opt}^R$, $\boldsymbol{p}_{w\_r\_opt}^R$), and initial object pose $\boldsymbol{x}_{o}^R$. Define the vector between the human wrist positions as $\boldsymbol{v}_{init} = \boldsymbol{p}_{w\_l}^R - \boldsymbol{p}_{w\_l}^R$ and the vector between the optimized human wrist positions as $\boldsymbol{v}_{opt} = \boldsymbol{p}_{w\_l\_opt}^R - \boldsymbol{p}_{w\_r\_opt}^R$, the normalized vectors can be written as $\boldsymbol{v}_{norm} = \boldsymbol{v} / ||\boldsymbol{v}||$. Then the rotation matrix between the initial vector and the optimized vector can be given as:
\begin{equation}
\boldsymbol{R}_{o} = \boldsymbol{I} + \sin{\theta} \cdot \boldsymbol{S}_r + (1 - \cos{\theta}) \cdot \boldsymbol{S}^2_r ,
\end{equation}
where $\boldsymbol{S}_r$ is the skew-symmetric matrix of the rotation vector $\boldsymbol{r} = \boldsymbol{v}_{init\_norm} \times \boldsymbol{v}_{opt\_norm}$, $\theta = \boldsymbol{v}_{init\_norm} \cdot \boldsymbol{v}_{opt\_norm}$ represents the rotation angle.

The orientation of the updated object pose can be generated based on $\boldsymbol{R}_{o}$ and $\boldsymbol{x}_{o}^R$. Since we have the vector between the initial left wrist position and the initial object position $\boldsymbol{v}_{w\_l}$, the vector between the optimized left wrist pose and the corresponding object pose is written as $\boldsymbol{v}_{l\_opt} = \boldsymbol{R}_{o} \cdot \boldsymbol{v}_{l\_init}$. Then the updated position of the object can be calculated based on the optimized left wrist position $\boldsymbol{p}_{w\_l\_opt}^R$ and the vector $\boldsymbol{v}_{l\_opt}$.

With the initial poses of the cobot end effectors $\boldsymbol{x}_{r\_l}^R$ and $\boldsymbol{x}_{r\_r}^R$, the corresponding vectors from the initial object pose to each end effector pose are represented as $\boldsymbol{v}_{r\_l}$ and $\boldsymbol{v}_{r\_r}$. The updated vectors can be calculated based on the rotation matrix $\boldsymbol{R}_{o}$. Hence, the updated positions of the bimanual cobot end effector are given as $\boldsymbol{p}_{r\_l\_opt}^R$ and $\boldsymbol{p}_{r\_r\_opt}^R$. Similarly, the orientations of the updated cobot poses are generated by multiplying the initial rotation with the rotation matrix $\boldsymbol{R}_{o}$. Finally, the updated poses $\boldsymbol{x}_{r\_l\_opt}^R$ and $\boldsymbol{x}_{r\_r\_opt}^R$ are given for the cobot end effector.

\begin{figure*}[!t]
\centering
\includegraphics[width=1\linewidth]{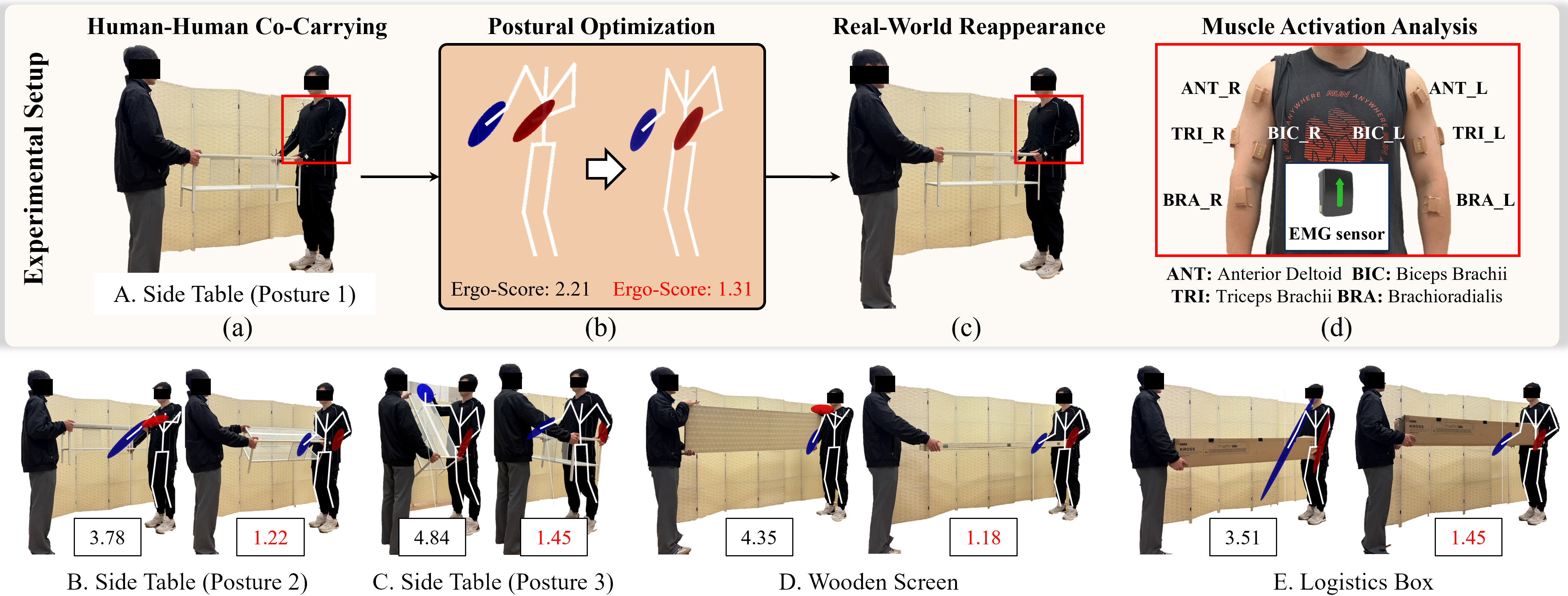} 
\caption{Experimental validation by human-human co-carrying of diverse objects. The experimental setup includes: (a) data collection of initial posture with the motion capture system, (b) postural optimization to generate a new posture, (c) real-world reappearance by the subject to mimic the optimized posture, (d) muscle activation analysis of target muscles before and after optimization within 5 $s$. Five demonstrations are conducted to co-carry the side table (A-C), wooden screen (D), and logistics box (E).}
\label{Fig.exp1}
\end{figure*}

\subsection{Controller}
The minimum jerk trajectory planning method is applied to generate the trajectories for the dual arms of the robot based on both initial and updated poses. Subsequently, an impedance control strategy is employed on our cobot, named CURI, to facilitate HRC. CURI is a dual-arm mobile manipulator equipped with two torque-controlled 7-DoF Franka Emika Panda arms, as depicted in Fig. \ref{Fig.cover}.

The dynamic behavior of each arm is modeled by the following equation:
\begin{equation}\label{robot dynamics}
\bm{M}(\bm{\hat{q}})\bm{\ddot{\hat{q}}} + \bm{C}(\bm{\hat{q}}, \bm{\dot{\hat{q}}})\bm{\dot{\hat{q}}} + \bm{g}(\bm{\hat{q}})
= \bm{\tau} + \bm{\tau}_{e} ,
\end{equation}
In this model, $\bm{\hat{q}}$ denotes the vector of robot joint angles, $\bm{M}(\bm{\hat{q}})$ represents the inertia matrix in joint space, $\bm{C}(\bm{\hat{q}}, \bm{\dot{\hat{q}}})$ refers to the Coriolis and centrifugal forces matrix, and $\bm{g}(\bm{\hat{q}})$ indicates the gravitational force vector. The terms $\bm{\tau}$ and $\bm{\tau}_{e} = \bm{J}_r^T \bm{f}_e$ signify the control input torques and the external torques in the joint space, respectively, where $\bm{J}_r$ is the Jacobian matrix of the robotic arm.

Leveraging the defined dynamics and interaction models \cite{li2024towards}, the reference joint torques $\bm{\tau}$ are computed as follows:
\begin{equation}
\label{control_torque}
\begin{aligned}
\bm{\tau} = \bm{J}_r^T\bm{u} + \bm{g}(\bm{\hat{q}}) +\bm{C}(\bm{\hat{q}}, \bm{\dot{\hat{q}}}) ,
\end{aligned}
\end{equation}
where $\bm{u}$ represents the optimal control force in Cartesian space based a model predictive impedance controller for bimanual collaboration, which is expressed as:
\begin{equation}
\label{cost_function}
\begin{aligned}
\min_{u_k} \mathcal{J}(\bm{u},\tilde{\bm{X}}_k) & = \min_{u_k} 
\sum_{k=0}^{N-1} \big( \| \bm{w}_k + \bm{\mathcal{K}}_{I,k}\tilde{\bm{X}}_k \|{_{\bm{Q}_I}^2}  \\
& +  \| \bm{v}_{k} +  
\bm{C}^T\bm{\mathcal{K}}_C\bm{C}\tilde{\bm{\mathcal{X}}}_k\|{_{\bm{Q}_C}^2} \\
& + \| \bm{s}_{k} + 
\bm{F}_{e,k} - \bm{\mathcal{K}}_{F,k}\tilde{\bm{F}}_k \|{_{\bm{Q}_F}^2} \\
& +  \| \bm{u}_{k} - \bm{w}_{k} - \bm{v}_{k} - \bm{s}_{k} \|{_{\bm{Q}_u}^2} \big) \\
\text{s.t.} \quad \dot{\bm{\mathcal{X}}}_{k+1} &= \bm{A}_{k} \bm{\mathcal{X}}_{k} + \bm{B}_{k} \bm{u}_{k} \\
& \| \bm{u}_{i,j,k} \| \leq \overline{\bm{u}}_{i,j} \\
& \|\bm{\mathcal{X}}_{i,j,k}\| \leq \overline{\bm{\mathcal{X}}}_{i,j} \\
\end{aligned}
\end{equation}
where $\bm{\mathcal{K}}_{I,k}\tilde{\bm{\mathcal{X}}}_k$ is used for constructing impedance behavior for each single arm. $\bm{C}^T\bm{\mathcal{K}}_C \bm{C} \tilde{\bm{\mathcal{X}}}_k$ is employed to establish collaborative behavior between dual-arms. Collaborative control proposed to coordinate the motion of dual arms in the presence of external disturbances and interaction modeling uncertainties. $\bm{C}$ is collaborative matrix, which defining the coordination constraints between the two arms. The term $\bm{F}_{e,k} - \bm{\mathcal{K}}_{F,k} \tilde{\bm{F}}_k$ is represents direct force feedback control. The subject $\dot{\bm{\mathcal{X}}}_{k+1} = \bm{A}_{k} \bm{\mathcal{X}}_{k} + \bm{B}_{k} \bm{u}_{k}$ represents the interaction model between the robot and the environment. Furthermore, the control force optimized by the model predictive impedance controller can also constrain the state of the robot, ensuring safe human-robot interaction in our co-carrying task.

\section{Experiments}
Two types of experiments were conducted to validate the proposed method. Initially, the postural optimization module was evaluated through human-human object co-carrying experiments. Subsequently, the overall collaborative framework was tested involving human-robot co-carrying tasks. 

\subsection{Human-Human Collaboration}
To validate the proposed optimization module, a series of human-human co-carrying tasks were conducted and the experimental setup is shown in Fig. \ref{Fig.exp1}. During these tasks, the skeletal movements of a participant were captured using a motion capture system. The recorded postures were then processed by the optimization module to compute updated postures, which participants subsequently replicated in real-world settings. The ergonomic score $s(\bm{q})$ was decreased from 2.21 to 1.31 in the example. The component of the maximum eigenvalue $\lambda^{\boldsymbol{M^F}}_{max}$ along load direction was increased to provide more load-bearing capacity. Muscle activation levels were monitored using EMG sensors to compare the differences between the initial and updated postures. For each posture, participants were instructed to maintain the position for 5 $s$, allowing for the quantification of muscle activation levels in four selected muscles on each body side.

The study utilized five demonstrations illustrated in Fig. \ref{Fig.exp1}. The first three demonstrations involved carrying a side table, each with a different posture undergoing optimization. The fourth demonstration featured a flat wooden screen, and the fifth involved a heavy logistics box weighing approximately 15 $kg$. Each demonstration was optimized to derive an optimized posture, which was then mimicked by the participants to measure and record muscle activations.

Table \ref{tab1} gives two examples of the average muscle activation levels with the initial posture ($init$) and the optimized posture ($opt$) during human-human collaboration (first two rows). The mean value and the maximum value of the eight muscle activations are calculated for further comparison. The optimized posture results in lower mean and maximum muscle activations. The mean value has been decreased 41.7$\%$ and 48$\%$ while the maximum value has been decreased 27.0$\%$ and 36.5$\%$ after postural optimization for trial B and D in Fig. \ref{Fig.exp1}, respectively.

\begin{figure*}[!t]
\centering
\includegraphics[width=1\linewidth]{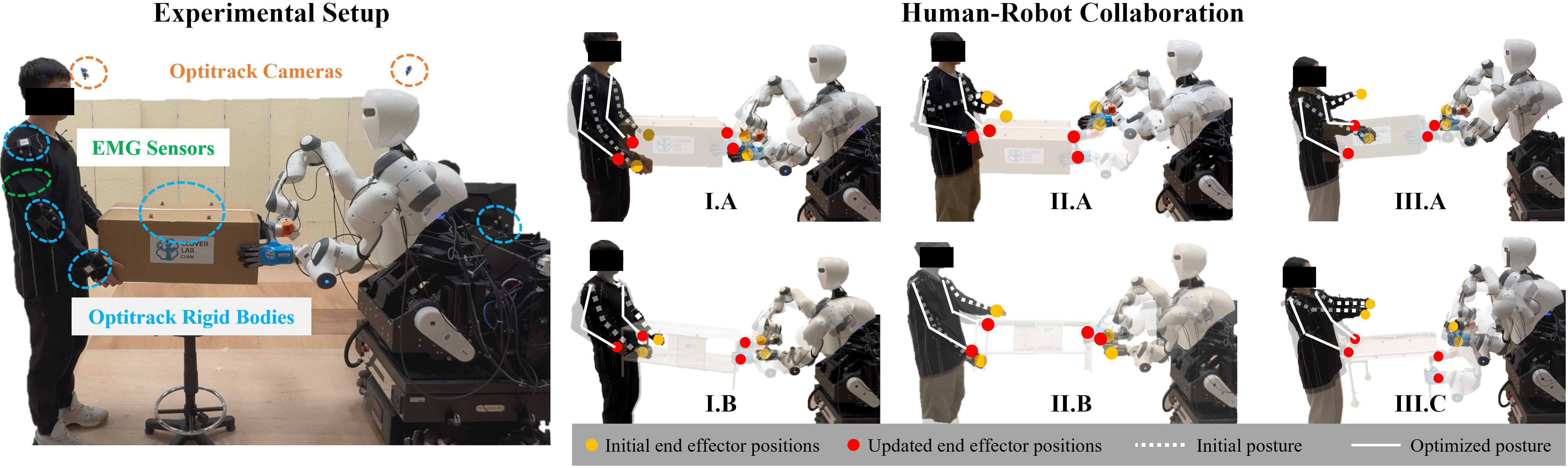} 
\caption{Real-world human-robot co-carrying with multiple objects and subjects. The experimental setup is introduced including Optitrack cameras and rigid bodies for human/robot/object motion capturing, EMG sensors for muscle activation recording. The snapshots of experiments are shown which include three subjects co-carry the logistics box (I.A-III.A), the side table (I.B-II.B), and the corner table (III.C) with cobot through diverse initial postures.}
\label{Fig.exp2} 
\end{figure*}

\begin{table*}[htbp]
\caption{average muscle activation comparison before and after postural optimization} 
\vspace{-5mm}
\begin{center}
\begin{tabular}{ccc c c c c c c c c c  }
\toprule
\multicolumn{1}{c}{\multirow{2}{*}{\textbf{\normalsize Trials}}}& \multicolumn{1}{c}{\multirow{2}{*}{\textbf{\normalsize State}}}& \multicolumn{8}{c}{\textbf{\normalsize Selected Muscles}} & \multicolumn{1}{c}{\multirow{2}{*}{\textbf{\normalsize Mean}}}& \multicolumn{1}{c}{\multirow{2}{*}{\textbf{\normalsize Max}}} \\
 &  & BIC\_L & TRI\_L & BRA\_L & ANT\_L & BIC\_R & TRI\_R & BRA\_R & ANT\_R \\

\midrule
\multirow{2}{*}{\textbf{Fig. 4.B} \textit{(hhc)}} & \textit{init} & 0.141 & 0.113 & 0.091 & 0.201 & 0.166 & 0.011 & 0.098 & 0.196 & 0.127 & 0.196 \\
& \textit{opt} & 0.056 & 0.117 & 0.133 & 0.035 & 0.067 & 0.011 & 0.143 & 0.030 & \textbf{0.074} & \textbf{0.143}\\

\addlinespace[0.4em]
\multirow{2}{*}{\textbf{Fig. 4.D} \textit{(hhc)}} & \textit{init} & 0.115 & 0.084 & 0.044 & 0.263 & 0.285 & 0.012 & 0.146 & 0.052 & 0.125 & 0.285 \\
& \textit{opt} & 0.036 & 0.181 & 0.078 & 0.036 & 0.035 & 0.013 & 0.113 & 0.026 & \textbf{0.065} & \textbf{0.181} \\
\midrule
\multirow{2}{*}{\textbf{Fig. 5.I.A} \textit{(hrc)}} & \textit{init} & 0.150 & 0.121 & 0.071 & 0.034 & 0.086 & 0.003 & 0.048 & 0.024 & 0.067 & 0.150 \\
& \textit{opt} & 0.106 & 0.123 & 0.071 & 0.025 & 0.073 & 0.002 & 0.050 & 0.016 & \textbf{0.058} & \textbf{0.123}\\

\addlinespace[0.4em]
\multirow{2}{*}{\textbf{Fig. 5.II.A} \textit{(hrc)}} & \textit{init} & 0.476 & 0.138 & 0.135 & 0.348 & 0.138 & 0.002 & 0.140 & 0.261 & 0.205 & 0.476 \\
& \textit{opt} & 0.348 & 0.151 & 0.203 & 0.081 & 0.098 & 0.002 & 0.140 & 0.020 & \textbf{0.131} & \textbf{0.348}\\

\addlinespace[0.4em]
\multirow{2}{*}{\textbf{Fig. 5.III.C} \textit{(hrc)}} & \textit{init} & 0.008 & 0.139 & 0.435 & 0.223 & 0.046 & 0.032 & 0.059 & 0.282 & 0.153 & 0.435 \\
& \textit{opt} & 0.022 & 0.055 & 0.169 & 0.037 & 0.042 & 0.019 & 0.085 & 0.061 & \textbf{0.061} & \textbf{0.169} \\
\bottomrule
\end{tabular}
\label{tab1}
\vspace{-4mm}
\end{center}
\end{table*}

\subsection{Human-Robot Collaboration}
Human-robot object co-carrying is then presented to verify the proposed framework. Three subjects are involved including two males and one female while the age averaged 28 years and body height averaged $1.74 \pm 0.14$ $m$. 

The experimental setup is given in Fig. \ref{Fig.exp2}. Joint information about the subject was collected in real-time using the Optitrack cameras and rigid bodies. Simultaneously, the object and the cobot were attached with markers to track the motions through this mo-cap system. Delsys EMG sensors were utilized to measure muscle activation levels, with deployment consistent with that in human-human collaboration. The lower limbs and the trunk of each subject are assumed to be constant before and after optimization. 

The experimental results of human-robot co-carrying are shown on the right side of Fig. \ref{Fig.exp2}. Three types of objects, including a logistics box, a side table, and a corner table, have been selected with weights of 3 $kg$, 4.5 $kg$, and 4 $kg$, respectively. Note that object I and III are both first occurrences for the subjects while object II has been selected in human-human collaboration. Initially, subjects I, II, and III will choose their preferred initial postures to grasp the object. For instance, as shown in Fig. \ref{Fig.exp2}, subject I carried the object by placing both hands on the bottom of the object to hold it in place (I.A); subject II placed one hand on the left top corner while the other on the right bottom corner of the object (II.A); Subject III put her right hand on the top while her left hand on the bottom to hold the object (III.A). The initial and updated human upper limb are represented with white solid lines and white dot lines, respectively. The initial and updated end effector positions of both the human and the robot are represented as orange solid circles and red solid circles, respectively. 6, 15, and 9 trials were done by subject I, II, and III respectively and a total of 30 trials were produced through human-robot collaboration.

\begin{figure}[!t]
\centering
\includegraphics[width=1\linewidth]{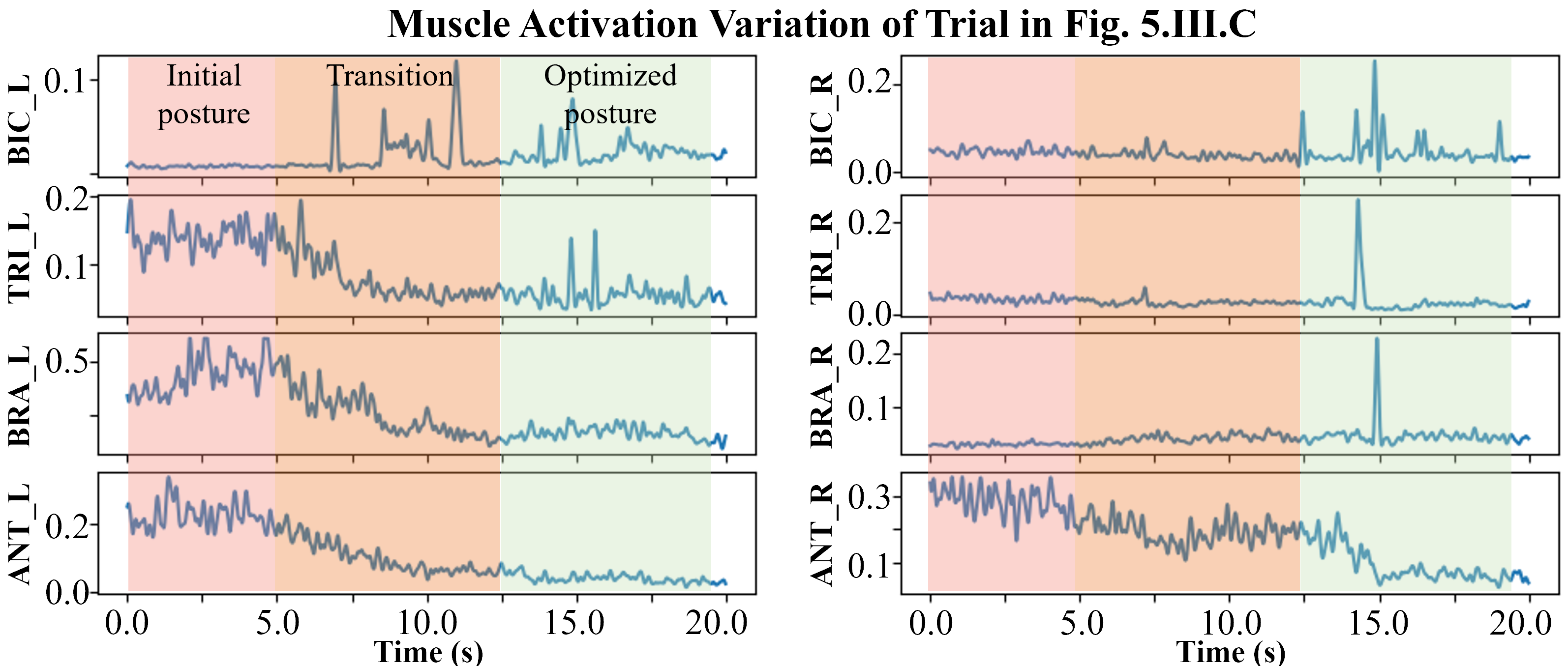} 
\caption{Muscle activation variation of target muscles during the trail III.C in Fig. \ref{Fig.exp2} is given as an example. The red area shows the muscle activation under the initial posture; The orange area is the transition by human-robot collaboration; The green area shows the result under the optimized posture.}
\label{Fig.exp_results} 
\end{figure}

Three experimental results (involving the three subjects) of the average muscle activation levels with the initial posture ($init$) and the optimized posture ($opt$) during human-robot collaboration (last three rows) in Table \ref{tab1}. An obvious drop in the mean activation value can be seen with the optimized postures for Trial I.A, II.A, and III.C, which are 13.4$\%$, 36.1$\%$, and 59.9$\%$, respectively. Meanwhile, the maximum activation values were also decreased with 18$\%$, 26.9$\%$, and 61.3$\%$, respectively. An example of the muscle activation variation of those target muscles is shown in Fig. \ref{Fig.exp_results}, which illustrates the EMG-based activation curves with the initial posture, transition, and optimized posture. It can be seen that significant drops of several muscle activations ($BRA\_L, TRI\_L, ANT\_L, ANT\_R$) appeared during the transition and stabilized at a low level with the optimized posture. Other muscles remain at their previous activation levels. The average drop of the mean and maximum activation is also given in Table \ref{tab2} throughout all trials for each subject to highlight the differences. Subject I and II (male) dropped $17.0 \pm 7.4\%$ and $36.6 \pm 15.5\%$ of their mean muscle activation after postural optimization while their maximum muscle activation dropped $8.9 \pm 5.4\%$ and $37.6 \pm 22.9\%$, respectively. Subject III (female) had the highest rate of decline with the mean value decreasing $39.2 \pm 18.3\%$ and the maximum value decreasing $42.2 \pm 15.8\%$. Consequently, adopting an optimized posture for carrying tasks significantly mitigates the risk of muscle fatigue when compared to pre-optimized postures. This innovative ergonomic strategy not only reduces the potential for physical strain but also substantially prolongs the period during which an individual can safely perform without compromising their health.

\begin{table}[!t]
\caption{average decrease of mean and maximum muscle activation} \vspace{-4mm}
\begin{center}
\begin{tabular}{cccc}
\toprule
\normalsize{\textbf{Subject}} & \normalsize{\textbf{Trial Quantity}} & \normalsize{\textbf{Mean} (\%)} &  \normalsize{\textbf{Max}} (\%) \\
\midrule
\small{I} & 6 & \small{17.0 $\pm$ 7.4} & \small{8.9 $\pm$ 5.4} \\
\small{II} & 15 & \small{36.6 $\pm$ 15.5} & \small{37.6 $\pm$ 22.9} \\
\small{III} & 9 & \small{39.2 $\pm$ 18.3} & \small{42.2 $\pm$ 15.8}  \\
\bottomrule
\end{tabular}
\label{tab2}
\vspace{-3mm}
\end{center}
\end{table}

\section{Conclusions}
In conclusion, this study proposes a novel upper-limb postural optimization method that significantly enhances both ergonomic safety and manipulative efficiency in bimanual human-robot co-carrying tasks. By optimizing joint angles of a simplified human skeletal model and implementing an MPIC for the CURI robot, our approach effectively guides human operators toward postures that minimize ergonomic risks while ensuring manipulability. Experimental validation involving diverse subjects and objects demonstrates significant improvements in human muscle conditions, as indicated by reduced muscle activation levels following optimization. The proposed method holds substantial potential for broader applications in enhancing human-robot collaboration across a wide range of operational scenarios.

\bibliographystyle{ieeetr}
\bibliography{ref_peo}

@article{mukherjee2022survey,
  title={A survey of robot learning strategies for human-robot collaboration in industrial settings},
  author={Mukherjee, Debasmita and Gupta, Kashish and Chang, Li Hsin and Najjaran, Homayoun},
  journal={Robotics and Computer-Integrated Manufacturing},
  volume={73},
  pages={102231},
  year={2022},
  publisher={Elsevier}
}

@inproceedings{evrard2009teaching,
  title={Teaching physical collaborative tasks: object-lifting case study with a humanoid},
  author={Evrard, Paul and Gribovskaya, Elena and Calinon, Sylvain and Billard, Aude and Kheddar, Abderrahmane},
  booktitle={2009 9th IEEE-RAS International Conference on Humanoid Robots},
  pages={399--404},
  year={2009},
  organization={IEEE}
}

@article{nemec2018human,
  title={Human robot cooperation with compliance adaptation along the motion trajectory},
  author={Nemec, Bojan and Likar, Nejc and Gams, Andrej and Ude, Ale{\v{s}}},
  journal={Autonomous robots},
  volume={42},
  pages={1023--1035},
  year={2018},
  publisher={Springer}
}

@article{da2010risk,
  title={Risk factors for work-related musculoskeletal disorders: a systematic review of recent longitudinal studies},
  author={Da Costa, Bruno R and Vieira, Edgar Ramos},
  journal={American journal of industrial medicine},
  volume={53},
  number={3},
  pages={285--323},
  year={2010},
  publisher={Wiley Online Library}
}

@inproceedings{van2020predicting,
  title={Predicting and optimizing ergonomics in physical human-robot cooperation tasks},
  author={van der Spaa, Linda and Gienger, Michael and Bates, Tamas and Kober, Jens},
  booktitle={2020 IEEE International Conference on Robotics and Automation (ICRA)},
  pages={1799--1805},
  year={2020},
  organization={IEEE}
}

@inproceedings{shafti2019real,
  title={Real-time robot-assisted ergonomics},
  author={Shafti, Ali and Ataka, Ahmad and Lazpita, B Urbistondo and Shiva, Ali and Wurdemann, Helge A and Althoefer, Kaspar},
  booktitle={2019 International Conference on Robotics and Automation (ICRA)},
  pages={1975--1981},
  year={2019},
  organization={IEEE}
}

@article{liao2023ergo,
  title={An Ergo-Interactive Framework for Human-Robot Collaboration Via Learning From Demonstration},
  author={Liao, Zhiwei and Lorenzini, Marta and Leonori, Mattia and Zhao, Fei and Jiang, Gedong and Ajoudani, Arash},
  journal={IEEE Robotics and Automation Letters},
  year={2023},
  publisher={IEEE}
}

@article{figueredo2020human,
  title={Human comfortability: Integrating ergonomics and muscular-informed metrics for manipulability analysis during human-robot collaboration},
  author={Figueredo, Luis FC and Aguiar, Rafael Castro and Chen, Lipeng and Chakrabarty, Samit and Dogar, Mehmet R and Cohn, Anthony G},
  journal={IEEE Robotics and Automation Letters},
  volume={6},
  number={2},
  pages={351--358},
  year={2020},
  publisher={IEEE}
}

@article{el2022virtual,
  title={A virtual element-based postural optimization method for improved ergonomics during human-robot collaboration},
  author={El Makrini, Ilias and Mathijssen, Glenn and Verhaegen, Sten and Verstraten, Tom and Vanderborght, Bram},
  journal={IEEE Transactions on Automation Science and Engineering},
  volume={19},
  number={3},
  pages={1772--1783},
  year={2022},
  publisher={IEEE}
}

@article{klopvcar2005kinematic,
  title={Kinematic model for determination of human arm reachable workspace},
  author={Klop{\v{c}}ar, Nives and Lenar{\v{c}}i{\v{c}}, Jadran},
  journal={Meccanica},
  volume={40},
  pages={203--219},
  year={2005},
  publisher={Springer}
}

@article{yoshikawa1985manipulability,
  title={Manipulability of robotic mechanisms},
  author={Yoshikawa, Tsuneo},
  journal={The international journal of Robotics Research},
  volume={4},
  number={2},
  pages={3--9},
  year={1985},
  publisher={Sage Publications Sage CA: Thousand Oaks, CA}
}

@article{hignett2000rapid,
  title={Rapid entire body assessment (REBA)},
  author={Hignett, Sue and McAtamney, Lynn},
  journal={Applied ergonomics},
  volume={31},
  number={2},
  pages={201--205},
  year={2000},
  publisher={Elsevier}
}

@article{al2021discrete,
  title={Discrete crow-inspired algorithms for traveling salesman problem},
  author={Al-Gaphari, Ghaleb H and Al-Amry, Rowaida and Al-Nuzaili, Afrah S},
  journal={Engineering Applications of Artificial Intelligence},
  volume={97},
  pages={104006},
  year={2021},
  publisher={Elsevier}
}

@article{bicchi1995mobility,
  title={On the mobility and manipulability of general multiple limb robots},
  author={Bicchi, Antonio and Melchiorri, Claudio and Balluchi, Daniele},
  journal={IEEE Transactions on Robotics and Automation},
  volume={11},
  number={2},
  pages={215--228},
  year={1995},
  publisher={IEEE}
}

@inproceedings{li2024towards,
  title={Towards Robo-Coach: Robot Interactive Stiffness/Position Adaptation for Human Strength and Conditioning Training},
  author={Li, Chenzui and Wu, Xi and Teng, Tao and Calinon, Sylvain and Chen, Fei},
  booktitle={2024 IEEE International Conference on Robotics and Automation (ICRA)},
  pages={860--866},
  year={2024},
  organization={IEEE}
}

@article{cardoso2021ergonomics,
  title={Ergonomics and human factors as a requirement to implement safer collaborative robotic workstations: A literature review},
  author={Cardoso, Andr{\'e} and Colim, Ana and Bicho, Estela and Braga, Ana Cristina and Menozzi, Marino and Arezes, Pedro},
  journal={Safety},
  volume={7},
  number={4},
  pages={71},
  year={2021},
  publisher={MDPI}
}

@article{raessa2020human,
  title={Human-in-the-loop robotic manipulation planning for collaborative assembly},
  author={Raessa, Mohamed and Chen, Jimmy Chi Yin and Wan, Weiwei and Harada, Kensuke},
  journal={IEEE Transactions on Automation Science and Engineering},
  volume={17},
  number={4},
  pages={1800--1813},
  year={2020},
  publisher={IEEE}
}

\vspace{11pt}

\end{document}